\documentclass[journal]{IEEEtran}
\usepackage{hyperref}
\usepackage{multirow}
\usepackage{multicol}
\usepackage{makecell}

\usepackage{cite}

\ifCLASSINFOpdf
  \usepackage[pdftex]{graphicx}
  \usepackage{graphicx}
  \usepackage{subfigure}
  \usepackage[]{caption2} 
\else
\fi
\usepackage{amsmath}
\begin{document}
%
\title{STGC-GNNs: A GNN-based traffic prediction framework with a spatial-temporal Granger causality graph}
%
%
%

\author{
		Silu He,
		Qinyao Luo,
        Ronghua Du,
		Ling Zhao,
		and~Haifeng Li, ~\IEEEmembership{Member,~IEEE}
\thanks{This paper is supported by the National Natural Science Foundation of China (41871364, 42271481) and Hunan Provincial Natural Science Foundation of China (2022JJ30698).}
\thanks{S. He, Q. Luo, R. Du, L. Zhao, H. Li are with School of Geosciences and Info-Physics, Central South University (Corresponding author H. Li, Email: lihaifeng@csu.edu.cn).}
}

%
%

\markboth{Journal of \LaTeX\ Class Files,~Vol.~14, No.~8, August~2015}%
{Shell \MakeLowercase{\textit{et al.}}: Bare Demo of IEEEtran.cls for IEEE Journals}
%



\maketitle

\begin{abstract}
The key to traffic prediction is to accurately depict the temporal dynamics of traffic flow traveling in a road network, so it is important to model the spatial dependence of the road network. The essence of spatial dependence is to accurately describe how traffic information transmission is affected by other nodes in the road network, and the GNN-based traffic prediction model, as a benchmark for traffic prediction, has become the most common method for the ability to model spatial dependence by transmitting traffic information with the message passing mechanism. However, existing methods model a local and static spatial dependence, which cannot transmit the global-dynamic traffic information (GDTi) required for long-term prediction. The challenge is the difficulty of detecting the precise transmission of GDTi due to the uncertainty of individual transport, especially for long-term transmission. In this paper, we propose a new hypothesis: GDTi behaves macroscopically as a transmitting causal relationship (TCR) underlying traffic flow, which remains stable under dynamic changing traffic flow. We further propose spatial-temporal Granger causality (STGC) to express TCR, which models global and dynamic spatial dependence. To model global transmission, we model the causal order and causal lag of TCR’s global transmission by a spatial-temporal alignment algorithm. To capture dynamic spatial dependence, we approximate the stable TCR underlying dynamic traffic flow by a Granger causality test. The experimental results on three backbone models show that using STGC to model the spatial dependence has better results than the original model for 45-min and 1 h long-term prediction.
\end{abstract}


%
\IEEEpeerreviewmaketitle

\section{Introduction}
\label{section1}
\IEEEPARstart{W}{ith} the growing development of Intelligent Transportation Systems (ITS), traffic prediction, which is an important function of ITS, has received increasing attention. Traffic prediction is considered a time-series prediction task that requires the prediction of traffic data in the future based on historical traffic data recorded by traffic sensors and includes traffic flow prediction, flow velocity prediction, and peak hour prediction. These can support decision-making for city management, traffic planning, and route optimization.

The basic assumption of traffic prediction is that a stable pattern is implied behind the traffic data. Such patterns can be discovered from historical data and used for future forecasting. A large number of traffic forecasting studies have emerged in recent years. Statistics-based methods were first applied, including historical averaging (HA)\cite{1,2}, vector autoregression (VAR)\cite{3}, the autoregressive integrated moving average (ARIMA) model\cite{4,5,6,7,8,9}, and the Kalman filtering model\cite{10}. With the development of deep learning, regression models such as feedforward neural networks (FFNs)\cite{11,12,13,14} and deep belief networks (DBNs) are used for traffic prediction. A recurrent neural network (RNN)\cite{15,16} and its variants long short-term memory (LSTM) and gated recurrent units (GRUs)\cite{17} have also been used to model the temporal dependence of traffic data. However, the above methods only model the temporal dependence and ignore the spatial dependence among different road network nodes. This is mentioned in \hyperlink{observation1}{Observation 1}. Existing views generally agree that traffic prediction is different from the ordinary time-series prediction task in that the spatial dependence constrained to the road network structure is more important in addition to capturing temporal patterns.

To capture spatial dependence, models based on convolutional neural networks (CNNs) have been used to model spatial dependence\cite{18}. However, such models are only applicable to Euclidean data and not non-Euclidean road networks. Therefore, graph neural networks (GNNs), which can represent discrete and irregular data, is used to model complex road networks and has become the benchmark model for traffic prediction. These methods extract a local connectivity between nodes from the static road network topology and model it as a spatial graph to input into the GNN structure and then transmit the traffic information between nodes through a message-passing mechanism such as the temporal graph convolutional network (T-GCN)\cite{19}, graph wavenet\cite{20}, or spatiotemporal graph convolutional network (STGCN)\cite{21}. Other methods such as STSGCN\cite{22}, STFGNN\cite{23}, and STGODE\cite{24} also consider the effect of spatial dependence on the time-axis, and a combination of spatial graphs and temporal graphs is input into the GNN structure.

The essence of spatial dependence is to accurately describe how traffic information is transmitted between road network nodes, in other words, how the transmission of traffic information is influenced by other nodes. However, the large uncertainty of microtransport behavior makes it inherently difficult to track and predict the traffic information transmitted in the road network, especially in long-term prediction, where the transmission of traffic information is a global and dynamic process. The local and static spatial dependence described by the spatial graph cannot express the global-dynamic traffic information (GDTi) transmitted among the road network nodes. Therefore, how to model the transmission of GDTi in the road network and predict the traffic information in the future with the power of GNN is the key problem in long-term prediction.

In this paper, we propose a transmitting causal relationship (TCR) hypothesis: the transmission of GDTi is very uncertain at the microlevel but behaves as a stable causal relationship underlying traffic flow transmission at the macrolevel. TCR can be approximated by the spatial-temporal Granger causality test in the long-term transmission of GDTi through long-term transmission. We further propose a GNN-based traffic prediction framework with a spatial-temporal Granger causality graph (STGC-GNNS) for prediction. Our contributions are as follows.

\begin{enumerate}
\item{We propose a spatial-temporal Granger causality to model TCR, which can express global and dynamic spatial dependence and capture the transmission of GDTi to perform long-term prediction tasks. Spatial-temporal Granger causality can be detected by a spatial-temporal alignment algorithm followed by a Granger causality test.}

\item{We propose a GNN-based traffic prediction framework with a spatial-temporal Granger causality graph (STGC-GNNS) to improve long-term prediction. This framework is compatible with all GNN-based traffic prediction models that use only a spatial graph to capture spatial dependence.}

\item{We conduct comparative performance experiments on a real-world dataset, and the results show that our method can outperform three backbone models on horizons of 45 min and 60 min. The visualization of the results shows that our method can improve the prediction of nodes with high prediction difficulty on all horizons such as intersectional, boundary, and distant nodes. This can further verify the validity of the spatial dependence we captured.}

\end{enumerate}

The rest of the paper is organized as follows. In Section \ref{section2}, existing traffic prediction methods are summarized and analyzed. Our hypothesis and the observations deducing it are described in Section \ref{section3}. Section \ref{section4} depicts our overall framework and detailed methods. We evaluate our work in Section \ref{section5} and conclude this paper in Section \ref{section7}. Several potential problems to solve in the future are discussed in Section \ref{section6}.


\section{Related work}
\label{section2}
The traffic prediction task aims to forecast future traffic data using historical traffic data and includes traffic flow prediction, flow velocity prediction, and peak hour prediction. It is an important part of Intelligent Transportation Systems (ITS), and existing traffic prediction methods can be classified into model-driven and data-driven types. Model-driven methods consider traffic prediction as a time-series forecasting task, aiming to model the temporal patterns inherent in historical data and use them for prediction. These can be classified into two categories: statistical methods and machine learning methods. Among them, statistical methods were first used in traffic flow forecasting by fitting parametric models from historical data through parametric methods such as linear regression for forecasting, including historical averaging (HA)\cite{1,2}, vector autoregression (VAR)\cite{3}, autoregressive integrated moving average (ARIMA) models\cite{4,5,6,7,8,9}, and Kalman filtering models\cite{10}. These models are based on the assumption of stationary and linear time series and can make simple and fast forecasts but are not sufficient to model the uncertainty in complex and dynamic traffic data. Machine learning methods fill this gap by automatically learning the patterns inherent in the data from sufficient historical data based on nonlinear assumptions. This includes including feature-based methods, Gaussian process-based methods, and state-space-based methods. Feature-based methods regress traffic data using human-engineered important traffic features\cite{25,26,27}, Gaussian process-based methods model the intrinsic features of traffic data for prediction by different kernel functions\cite{28,29,30}, and state space-based methods consider the traffic data generation process as a hidden Markov process and thus use the state space model to model the traffic system\cite{31,32,33,34,35,36,37,38,39,40,41,42,43,44,45}. Machine learning methods are able to handle high-dimensional data and capture the complex nonlinearity inherent in the data.

With the development of deep learning, as a powerful approximator, deep neural networks can learn implicit patterns from large amounts of data. These are called data-driven methods and have achieved good performance in traffic prediction tasks. Data-driven traffic prediction methods can be classified into two categories: one for considering only temporal dependence and the other for considering both temporal and spatial dependence. The temporal-dependence-only approaches use CNN\cite{46}, RNN\cite{15,16}, and its variants\cite{17} to model the temporal correlation in traffic data. These methods can capture temporal characteristics such as the periodicity and trend of traffic flow data but ignore the spatial dependence. The methods that capture both temporal and spatial dependence consider the interactions between road network nodes and can be divided into CNN-based methods and GNN-based methods based on the representation of the relationships between nodes. CNN-based methods divide road networks into regular grids and ignore the natural discrete, irregular structure of road networks\cite{18}; GNN-based methods take this into account, are capable of characterizing non-Euclidean structures, and have become a benchmark model of traffic prediction. According to the GNN framework used, these approaches can be classified as GCN-based methods, GAT-based methods, and GAE-based methods. GCN-based methods model spatial dependencies using convolutional operations in the spatial or spectral domain\cite{19,20,21,22,23,24,47,48,49,50}, GAT-based methods use attention mechanisms to learn the weights of other traffic nodes\cite{51,52,53,54,55}, and GAE-based methods use a self-encoder structure to encode node representations\cite{56}.

GNN-based methods take the predefined graph as input to model the spatial dependence of complex road networks. The existing methods extract a local connectivity between nodes from the static road network topology and express it as a spatial graph such as a traffic network topology graph or spatial distance weighting graph. The former adjacency matrix sets the entry value to 1 if road Sections $i$ and $j$ are topologically adjacent, while the latter preserves the connections between pairs of nodes that are closer by using a Gaussian filter. The entry value is closer to 1 if road sections $i$ and $j$ are spatially closer. Most methods input spatial graphs directly into GNN structures such as temporal graph convolutional networks (T-GCNs)\cite{19}, graph graphs\cite{20}, and spatiotemporal graph convolutional networks (STGCNs)\cite{21}. Another class of methods considers the similarity of time series between nodes on the basis of a spatial graph and combines a temporal graph with it to input. STSGCN\cite{22} considers spatial dependencies to be invariant on the time axis and thus models spatial-temporal correlation by connecting individual spatial graphs of adjacent time steps into one graph as input. STFGNN\cite{23} and STGODE\cite{24}, on the other hand, assume that spatial dependencies can lead to correlations on time series and therefore generate temporal graphs using DTW to compute correlations between time series, with the former merging the temporal graph with the spatial graph and the latter inputting the temporal graph and spatial graph into the prediction module separately.

Based on the above, the modeling of spatial dependence is particularly important for GNN-based methods. The essence of spatial dependence is to accurately describe the transmission of traffic information on road networks, which is very difficult in traffic prediction on complex road networks, especially in long-term prediction. The local and static spatial dependence modeled by existing methods cannot describe the state of global diffusion of traffic information after long-range and dynamic transmission, i.e., a spatial graph cannot model the transmission of GDTi in the road network. Modeling the transmission of GDTi in the road network and predicting the traffic information with GNN is the key problem in long-term prediction.

To capture the transmission of GDTi, we propose a TCR hypothesis and further express the TCR using spatial-temporal Granger causality to achieve the modeling of global and dynamic spatial dependence.

\section{Transmitting Causal Relationship (TCR) Hypothesis}
\label{section3}
The modeling of spatial dependence essentially captures the transmission of traffic information between different nodes in the road network. For the long-term traffic prediction task, the traffic information is difficult to capture accurately because it is global and dynamic after long-term travel. The global property describes that the distance spanned by this transmission is long-range from a spatial perspective, while the dynamic property describes its uncertainty and complexity from a temporal perspective. To model the spatial transmission of GDTi as accurately as possible, we propose a hypothesis in this section, the transmitting causal relationship (TCR) hypothesis, which argues that the key to long-term prediction lies in modeling the global and dynamic transmission of traffic information in the road network, which shows high uncertainty and complexity at the micro level but behaves as a stable causal relationship at the macro level.

Since the observed traffic data in the real world are a consequence of the combination of dynamic changes in traffic information and complex topological constraints of the road network and the transmission of traffic information is not independent of the time and space, three dependencies can be derived, which makes traffic prediction a challenging task. In this section, to make our hypothesis and the subsequent expression of our method clearer, we first redefine the three dependencies and lags derived. We also describe three observations on real-world transportation systems, which can support the motivations and hypothesis. Note that when defining spatial dependence, since we want to capture a transmitted spatial dependence that is directed, the definitions of the target-node and source-node are derived, which may be novel for the traffic prediction domain.

\subsection{Definition of different dependencies}
\emph{(1)Temporal dependence}

\begin{figure*}[htp]
\centering
    \begin{minipage}[t]{0.3\textwidth}
        \centering
        \includegraphics[width= \linewidth]{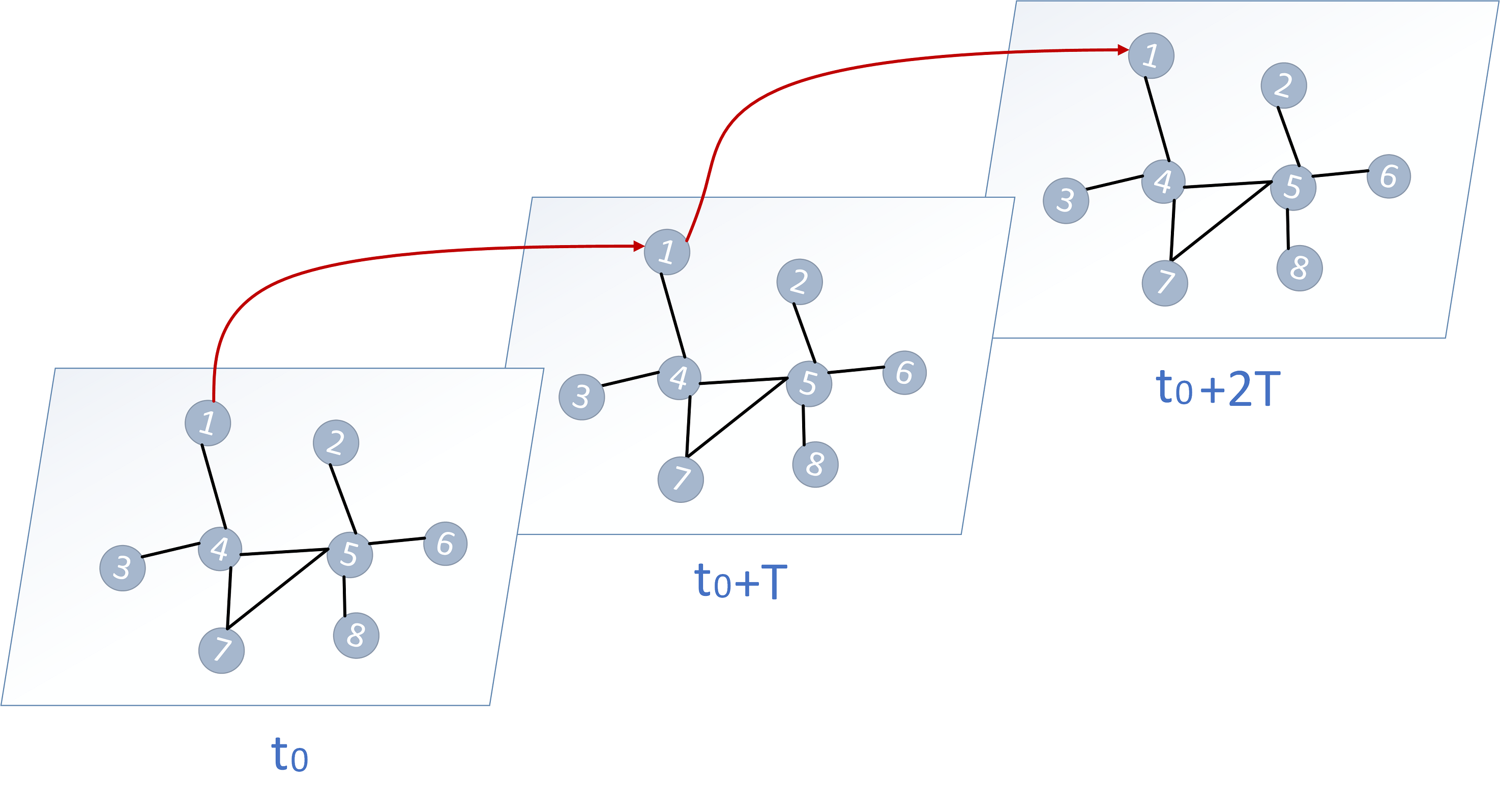}
        \caption{Temporal dependence in traffic data. Red arrow connects two nodes that are temporally dependent.}
        \label{fig1}
    \end{minipage}
    \begin{minipage}[t]{0.3\textwidth}
        \centering
        \includegraphics[width=\linewidth]{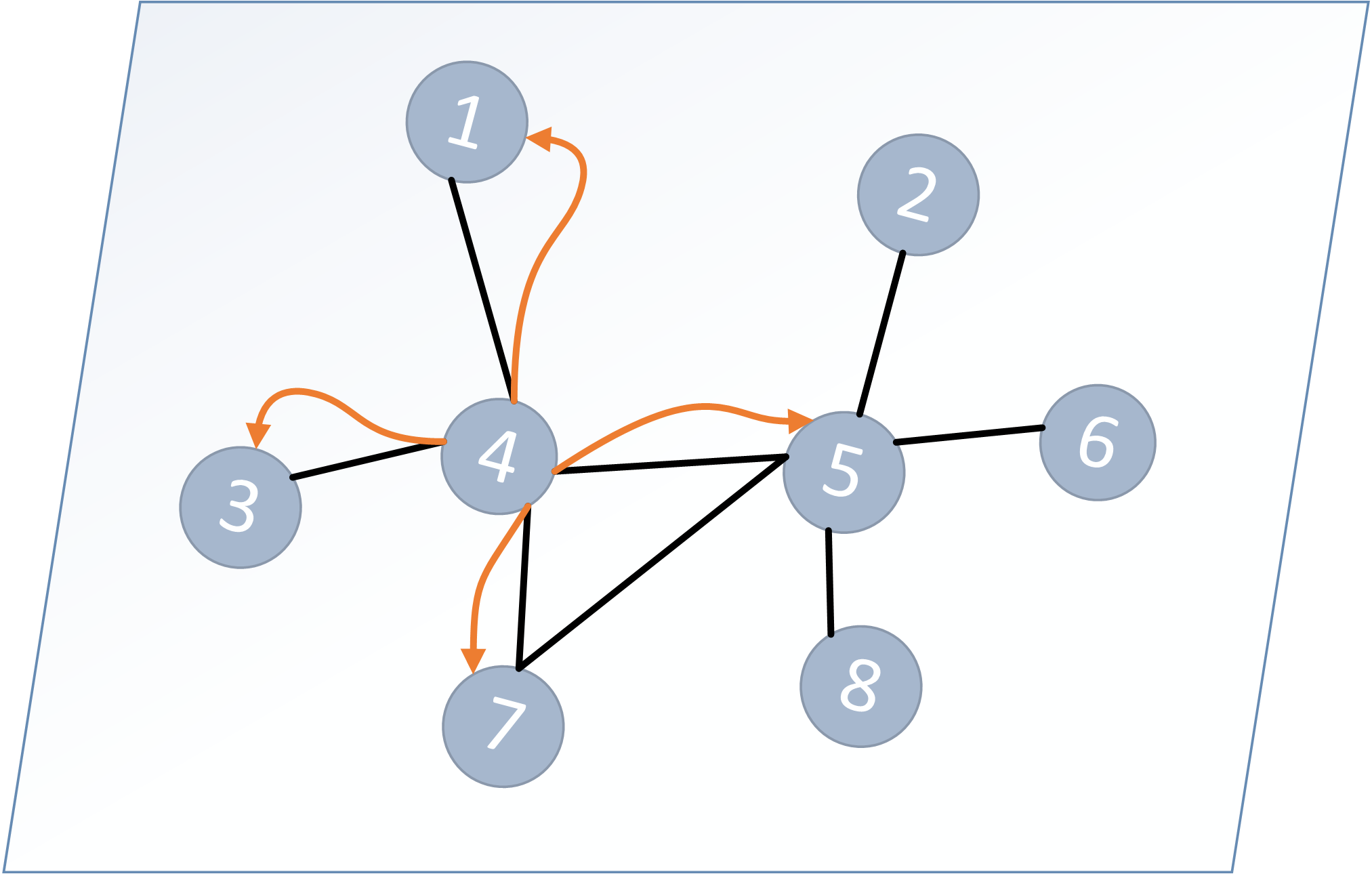}
        \caption{Spatial dependence in road network. Orange arrow connects two nodes that are spatially dependent.}
        \label{fig2}
        \end{minipage}
    \begin{minipage}[t]{0.3\textwidth}
        \centering
        \includegraphics[width=\linewidth]{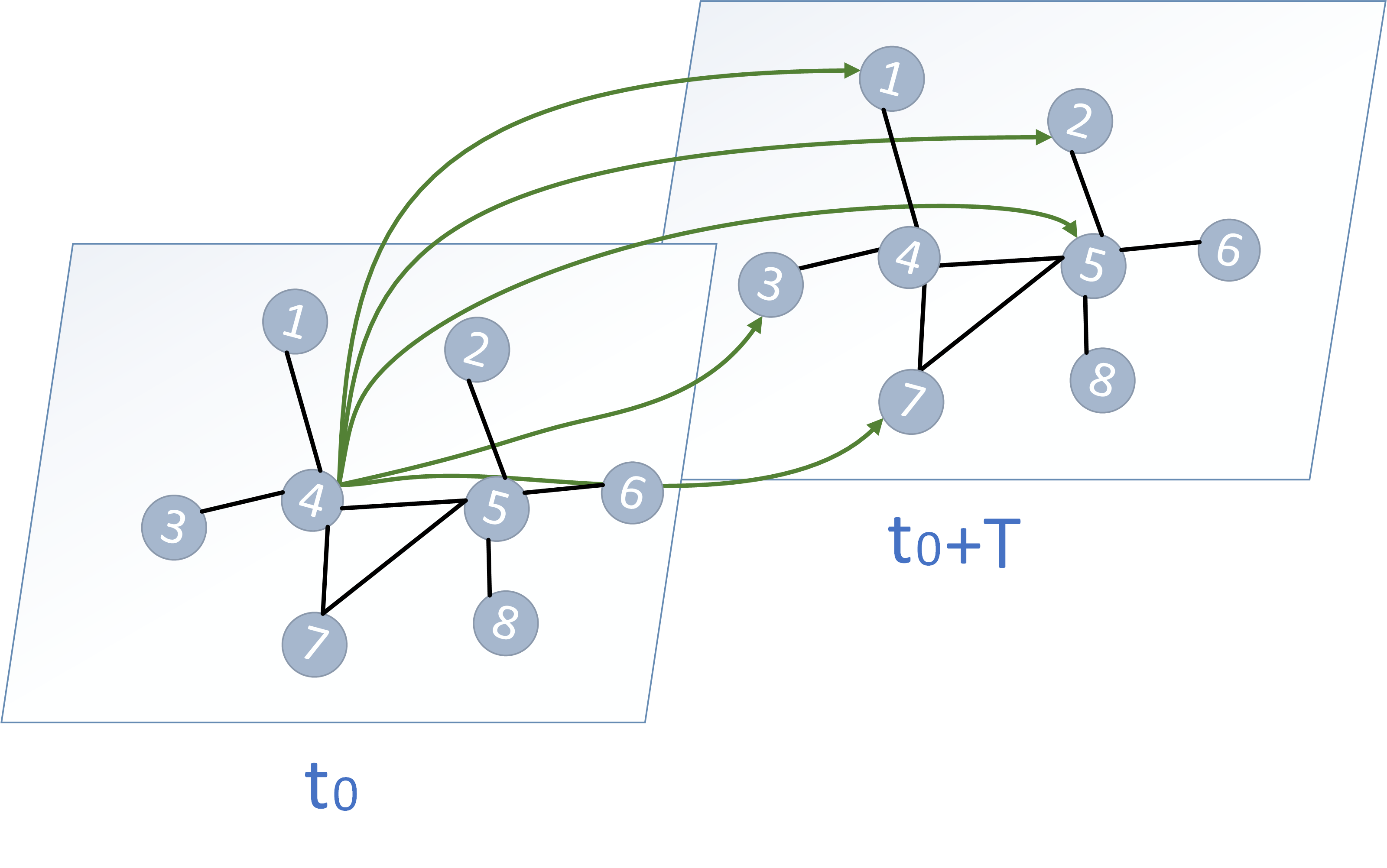}
        \caption{Spatial-temporal dependence in road network. Green arrow connects two nodes that are spatiotemporally dependent.}
        \label{fig3}
        \end{minipage}
\end{figure*}

\textbf{Definition 1} \emph{(Temporal dependence). Temporal dependence is the effect that the traffic information at the current timestep has on the traffic information at future timesteps for the same node.}

\textbf{Definition 2} \emph{(Temporal lag). Temporal lag is the interval of time to transmit traffic information between two timesteps, which are temporally dependent.}

\emph{(2)Spatial dependence}

\textbf{Definition 3} \emph{(Spatial dependence). Spatial dependence is the effect that the traffic information of one node has on the traffic information of another node at the current time step.}

\textbf{Definition 4} \emph{(Spatial lag). Spatial lag is the distance to transmit traffic information between two nodes, which are spatially dependent at one timestep.}

\textbf{Definition 5} \emph{(Target-node and Source-node). For a pair of nodes that are spatially dependent, we call the node that releases the traffic information the source-node and the node that receives the traffic information the target-node. As illustrated in Figure \ref{fig2}, node no. 4 is the source-node, and nodes no. 1, 2, 3, 5, and 7 are target-nodes.}

\emph{(3)Spatial-temporal dependence}

\textbf{Definition 6} \emph{(Spatial-temporal dependence). Spatial-temporal dependence is the effect that the traffic information of one node at the current time step has on the traffic information of another node at the future time step.}

\textbf{Definition 7} \emph{(Spatial-temporal lag). Spatial-temporal lag is the temporal lag consumed by transmitting traffic information between different nodes, which is spatial-temporal dependent.}

\subsection{Three observations on a real-world traffic system}
Similar to what was mentioned in Section \ref{section1}, existing methods ignore some important observations in real-world traffic systems. In this section, we will depict those observations that can help understand our methods.

\emph{(1)Observation 1: Power of spatial dependence}
\label{observation1}

\begin{figure*}[htp]
\centering  
\subfigure[]{
\label{fig4a}
\includegraphics[width=0.4\textwidth]{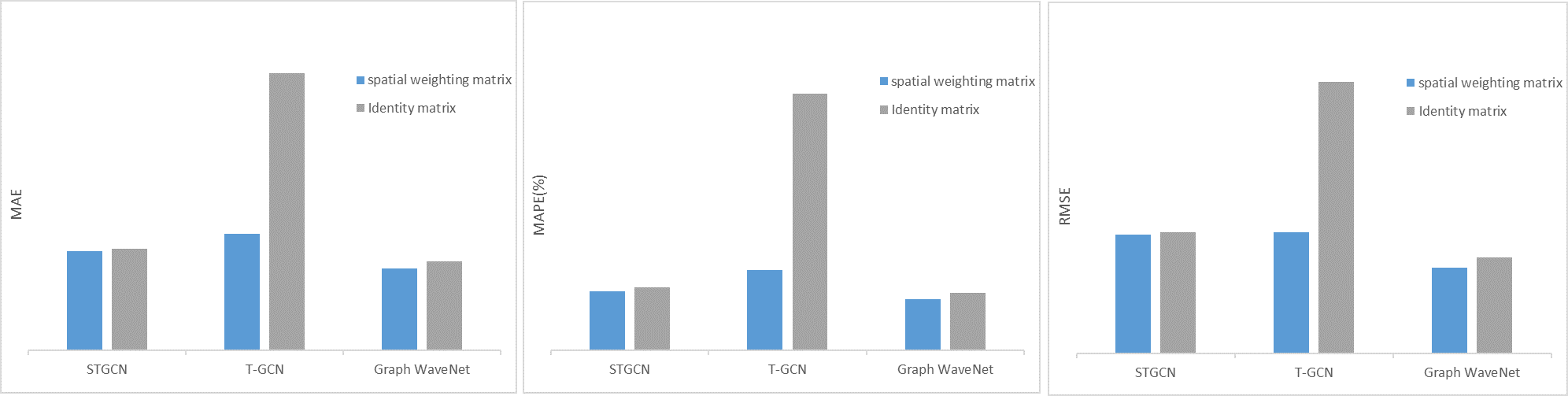}}
\subfigure[]{
\label{fig4b}
\includegraphics[width=0.4\textwidth]{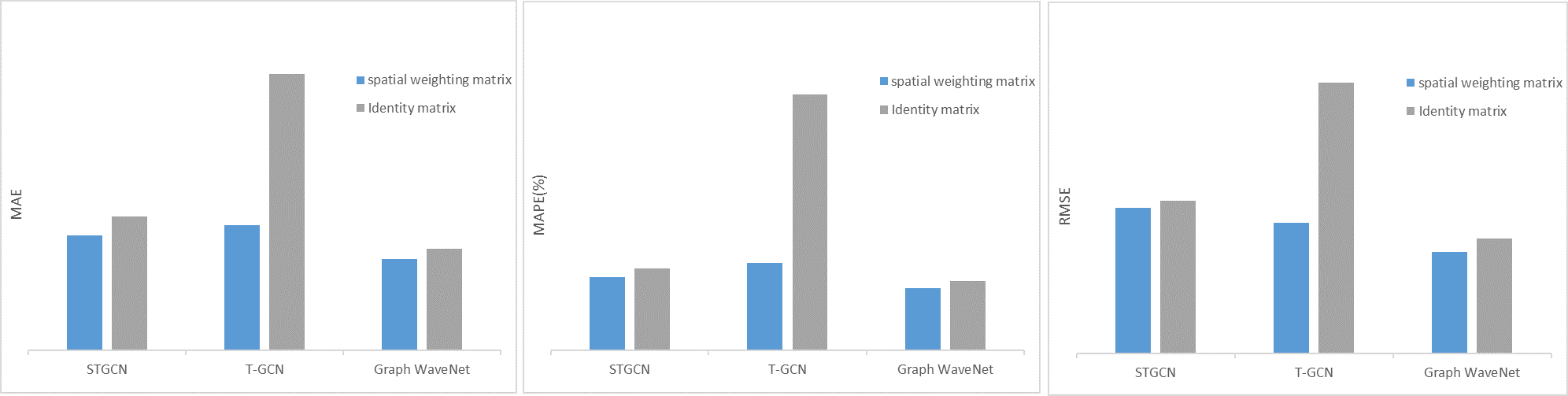}}
\subfigure[]{
\label{fig4c}
\includegraphics[width=0.4\textwidth]{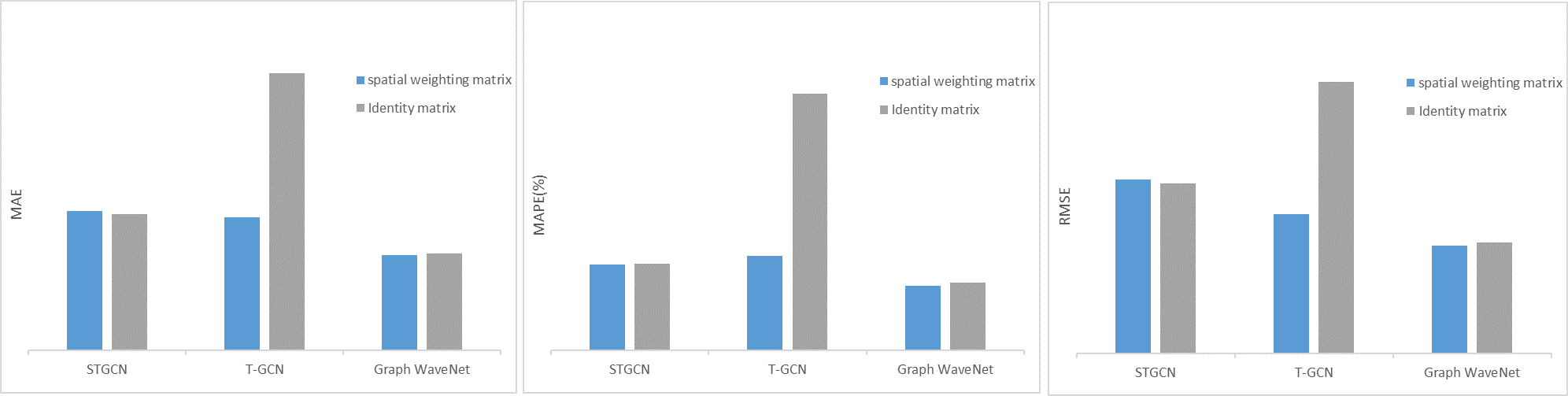}}
\subfigure[]{
\label{fig4d}
\includegraphics[width=0.4\textwidth]{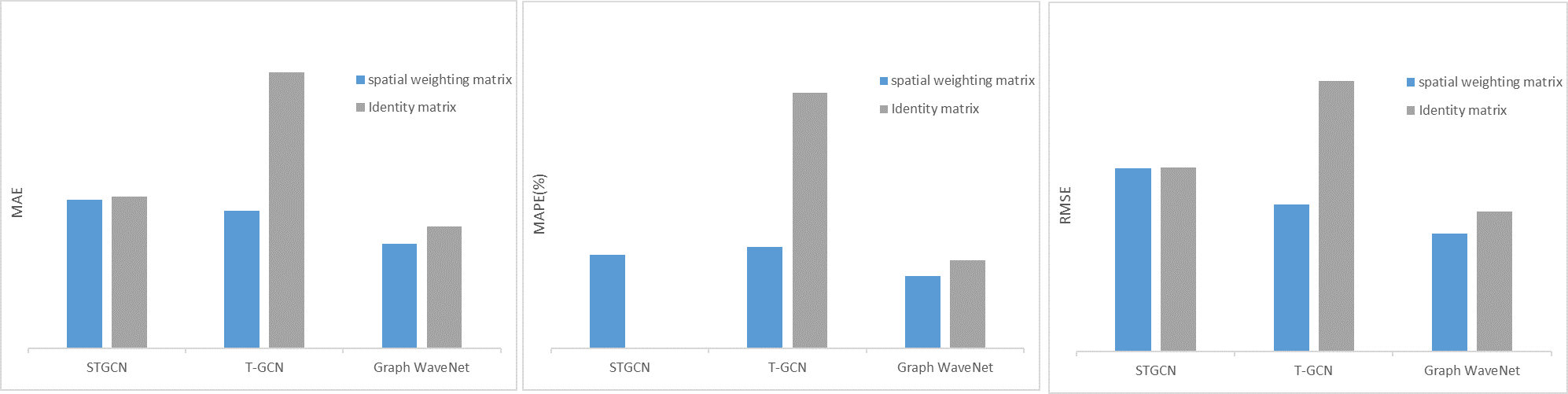}}
\caption{Comparison of prediction result of spatial graph as input with self-connected graph as input. (a) Horizon of 15 min. (b) Horizon of 30 min. (c) Horizon of 45 min. (d) Horizon of 60 min.}
\label{fig4}
\end{figure*}

The nodes in the road network are not completely independent but interact with others with the movement of transport, which is manifested as the transmission of traffic information and will lead to spatial dependence between interacting nodes. In the GNN-based model, the predictions of target-nodes can be improved if the spatial dependence can be modeled correctly because the information of other nodes is correctly introduced.

We observe that for the GNN-based model, since the spatial dependence is modeled by the input graph, it is possible to determine whether the input graph contains a valid spatial dependence by comparing its prediction result with the result when only self-connected inputs are used. Figure \ref{fig4} shows the prediction results of the model on the METR\_LA dataset when the input graph is a spatial graph (which is formulated as a spatial weighting matrix) and a self-connected graph (which is formulated as an identity matrix).

Figure \ref{fig4} shows that in the majority of cases, the prediction results of the spatial weighting matrix input outperform those of the identity matrix input, which indicates that it is necessary and effective to consider the spatial dependence. However, it can also be seen that on the horizon of 45 min, the prediction results of the STGCN with spatial graph input are instead worse than those with self-connected graph input, which indicates that when the captured spatial dependence is wrong or redundant, it may lead to worse performance compared to using the traffic information only from itself. We call the reason behind this phenomenon the power of spatial dependence.

Therefore, for the GNN-based traffic prediction model, the construction of input graphs modeling spatial dependence directly affects the prediction effect of the model even if all other settings of the model are kept consistent. This is exactly the power of spatial dependence, which must be taken into consideration when using GNN-based traffic prediction models.

\emph{(2)Observation 2: Spatial-temporal entanglement of traffic information transmission}
\label{observation2}

\begin{figure}[!t]
\centering
\includegraphics[width=2.5in]{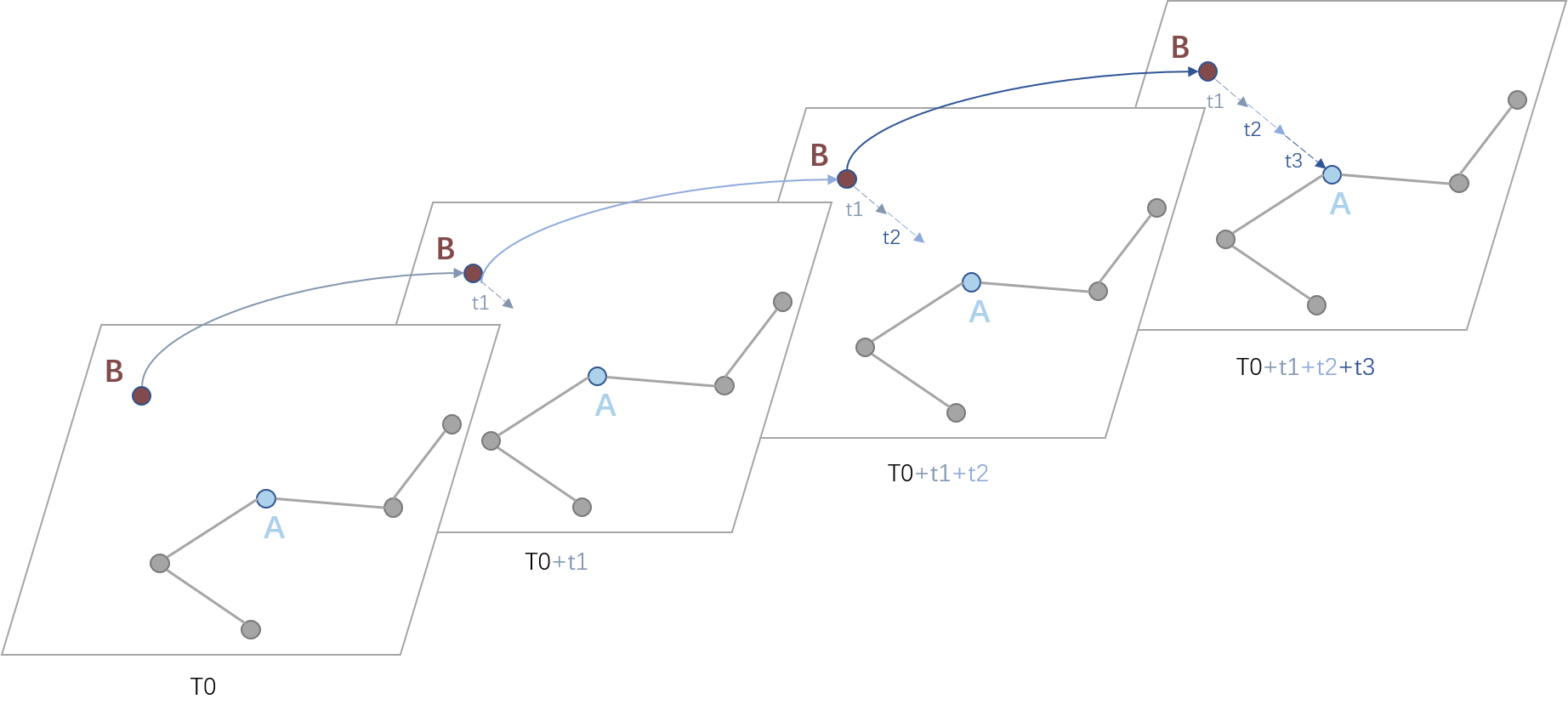}
\caption{Spatial-temporal entanglement of traffic information transmission. Arrow connecting different timesteps indicates transmission along time axis, while that connecting different nodes indicates transmission of this part of information along space axis.}
\label{fig5}
\end{figure}

In real-world transportation systems, the flow of traffic can be considered as a process of traffic information transmitting across space and time. This observation indicates that the spatial-temporal entanglement of traffic information transmission for the traveling distance and the corresponding time consumed are not independent. We depict this observation in Figure \ref{fig5}.

The observation of the spatial-temporal entanglement of traffic information transmission indicates that if we want to capture the spatial dependence of road networks from traffic information under global transmission, we must consider the spatial-temporal lag of traffic information transmission across space. We considered this in \hyperlink{algorithm1}{Algorithm 1}.

\emph{(3)Observation 3: Capability boundary of spatial distance}
\label{observation3}

\begin{figure}[!t]
\centering  
\subfigure[]{
\label{fig6a}
\includegraphics[width=0.4\textwidth]{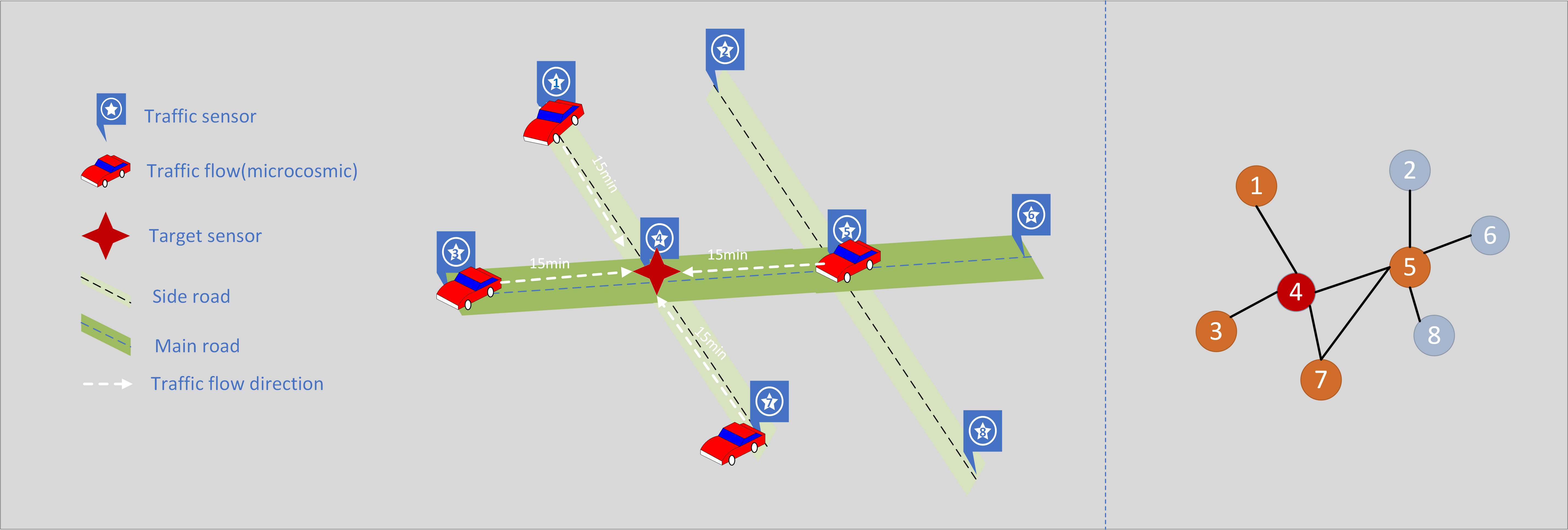}}
\subfigure[]{
\label{fig6b}
\includegraphics[width=0.4\textwidth]{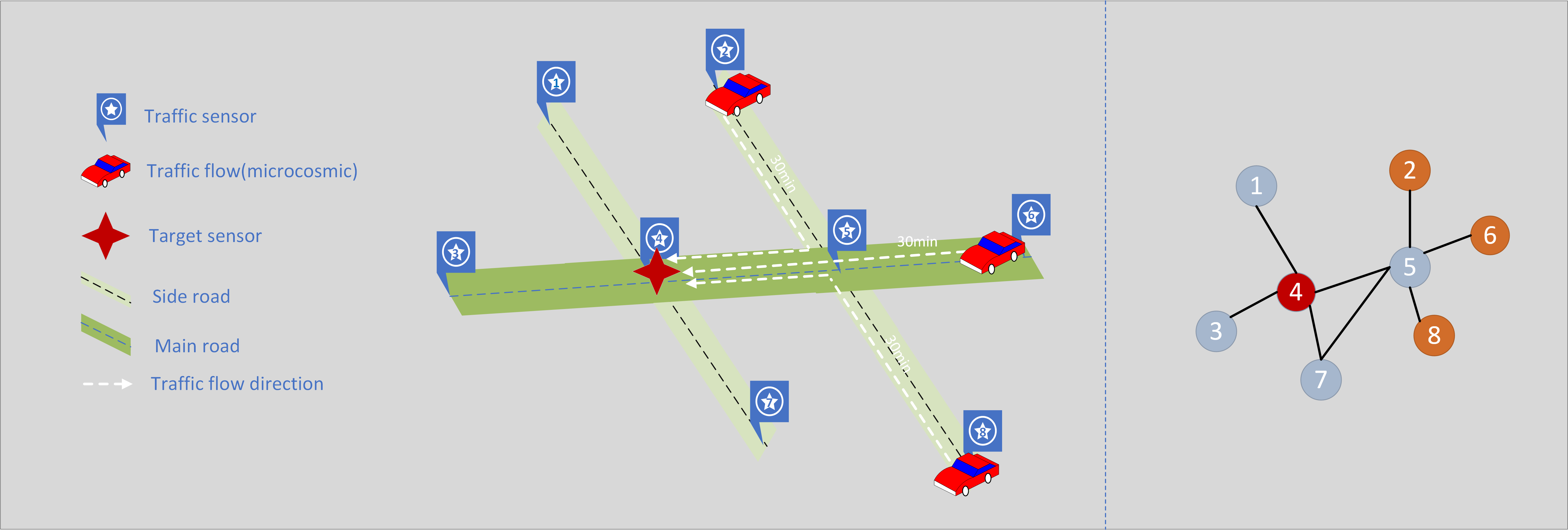}}
\caption{Simulating the transmission of traffic information on different prediction horizons. Red node is target-node, and orange node releases traffic information to target-node. (a) Horizon of 15 min. (b) Horizon of 30 min.}
\label{fig6}
\end{figure}

The existing methods generally consider the spatial distribution of road network nodes as an important indicator to measure spatial dependence, so they often use traffic network topology graphs or spatial distance weighting graphs as spatial graphs to input into GNN structures, which essentially model a local and static spatial dependence, i.e., the spatial dependency obtained from a fixed distribution of nodes in a certain road network. However, when traffic information is long-term transmitted, it will exhibit the property of being global in space and dynamic in time. We refer to this kind of traffic information as global-dynamic traffic information (GDTi). The key to long-term prediction is to model the transmission of GDTi. Therefore, it cannot be described by a local and static spatial dependence.

The simulation process of traffic flow for the same road network on horizons of 15 min and 30 min is illustrated in Figure \ref{fig6}. Assume that the time required for traffic information to be transmitted between two adjacent intersections in this traffic system is 15 min. Then, when we predict the traffic characteristics (e.g., flow rate and flow speed) of the target-node after 15 minutes, we only need to trace the traffic characteristics of the 1st-order neighbors at the current moment because this part of the traffic information will consume 15 minutes to transmit to the target-node. Similarly, when predicting the short-term traffic characteristics after 30 minutes, we need to trace the traffic characteristics of the 2nd-order neighbors.

It can be inferred that the nodes to be tracked will be farther away from the target-node on the prediction horizon of 45 minutes and longer, and therefore the transmission of GDTi with a traveling time of 45 minutes needs to be captured. This is difficult to capture in the existing spatial graph.

\subsection{TCR Hypothesis}
\label{tcr}

To capture the long-term transmission of GDTi in the road network as accurately as possible, we propose the TCR hypothesis and further hypothesize the properties of TCR based on the above observations, which can derive our method. According to \hyperlink{observation1}{Observation 1} and \hyperlink{observation3}{Observation 3}, we suggest that the spatial dependence for long-term prediction can be expressed by the transmission of GDTi, specifically as a kind of transmitting causal relationship (TCR), so we try to capture this TCR and model it directly as a graph input to the GNN structure. According to Observation 2, we consider the spatial-temporal entanglement of the transmission of GDTi and design the spatial-temporal alignment algorithm in the process of capturing the TCR.

\textbf{TCR Hypothesis. }\emph{The transmission of Global-Dynamic Traffic information (GDTi) appears as a stable causal relationship between nodes underlying transmitting traffic information, which is referred to as Transmitting Causal Relationship (TCR).}

\section{Methodology}
\label{section4}

\subsection{Overview}
\begin{figure*}[!t]
\centering
\includegraphics[width=5in]{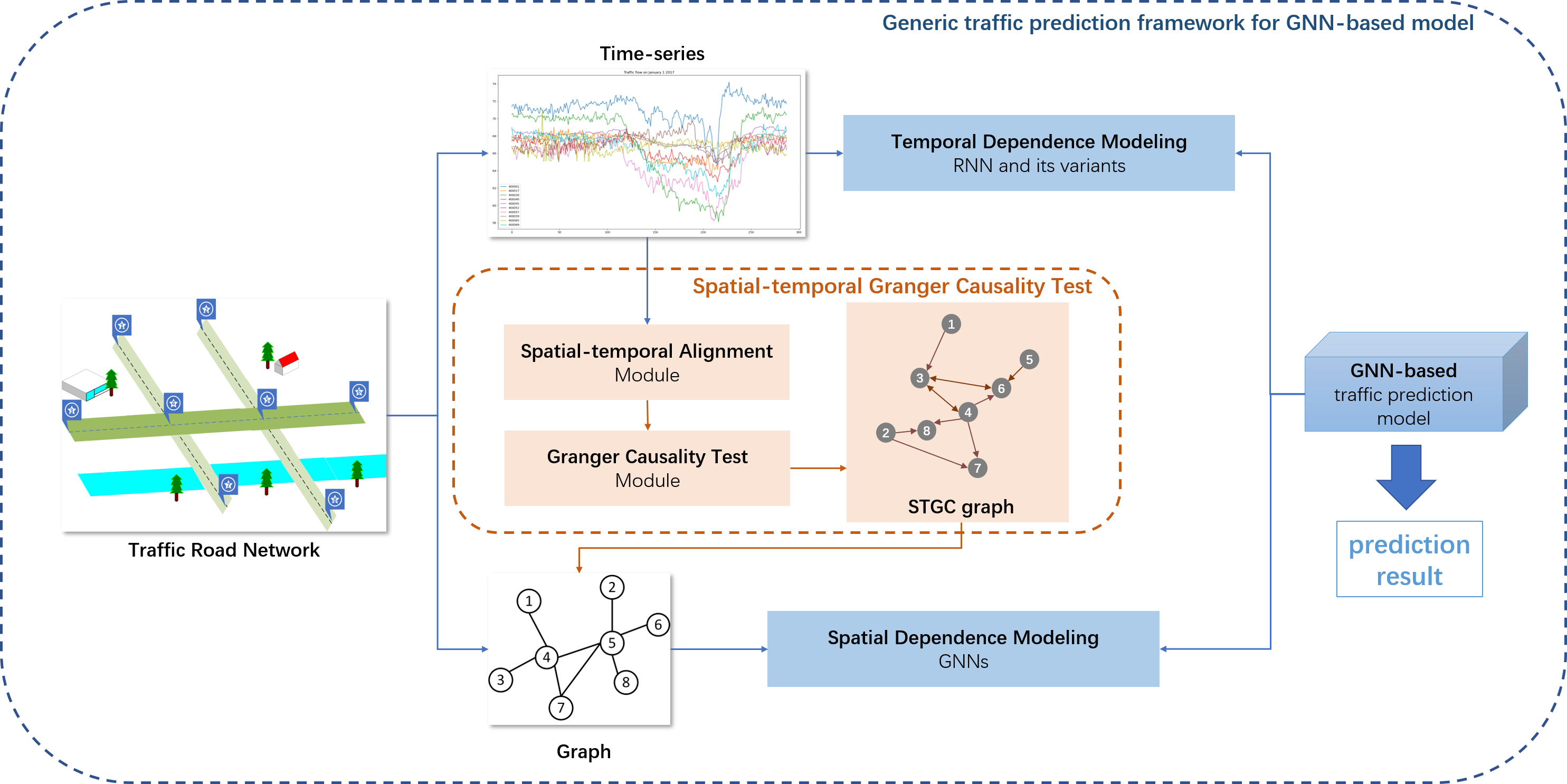}
\caption{Overview. Spatial-temporal Granger causality test is conducted to obtain STGC graph to input to GNN-based model for prediction.}
\label{fig7}
\end{figure*}

As illustrated in Figure \ref{fig7}, GNN-based traffic prediction models can always be formalized as a component that captures temporal dependence and a GNN that captures spatial dependence. However, existing spatial graphs describe a local and static spatial dependence that may fail to make long-term predictions. We propose a spatial-temporal Granger causality test method that captures global and dynamic spatial dependence and models it as a spatial-temporal Granger causality graph (STGC graph) input to the framework.

\subsection{Spatial-temporal Granger causality graph (STGC graph)}

In this section, we further propose a spatial-temporal Granger causality test to approximate the causal relationship underlying traffic road networks based on the TCR hypothesis to capture global and dynamic spatial dependence. The spatial-temporal Granger causality test is further divided into two modules: the spatial-temporal alignment module and the Granger causality test module. The former aligns the release and receive time of TCR’s global transmission, so the cause-node and effect-node contain the same part of GDTi at the same timestep, making transmission easier to detect; the latter further tests which two nodes the global transmission occurs from the aligned time-series, thus capturing the stable TCR under dynamic traffic flow.

\emph{(1)Spatial-temporal alignment}

According to the \hyperlink{tcr}{TCR hypothesis}, the transmission of GDTi can be regarded as the flow of causal effect from the cause-node to the effect-node, i.e., the mechanism of TCR. According to \hyperlink{observation1}{Observation 1} and \hyperlink{observation2}{Observation 2}, TCRs can be derived as having two significant properties.

\begin{itemize}
\item{\textbf{Causal lag. }The effect produced by the cause-node does not have an immediate impact on the effect-node. The occurrence of the cause and the effect has a time lag, which is expressed as a spatial-temporal lag between the cause-node and effect-node in the traffic road network.}
\item{\textbf{Causal order. }The effect always occurs or is observed after the cause, which is manifested as the effect-node’s receiving GDTi always occurring after the cause-node’s releasing GDTi.}
\end{itemize}

Therefore, we designed a spatial-temporal alignment algorithm, which can be divided into two steps.
\begin{enumerate}
    \item According to the causal lag, the time consumed by the TCR to transfer from the cause-node to the effect-node is calculated, i.e., the spatial-temporal lag $s$ between them.
    \item According to the causal order, we shift the time series of the cause-node by $s$ timesteps along the time axis to align the time when the TCR is generated on the cause-node and the time it impacts on the effect-node, thus ensuring that the causal mechanism can be inferred directly from the observed time series, which is the outcome of this causal mechanism.
\end{enumerate}

\begin{figure}[!t]
\centering
\includegraphics[width=2.5in]{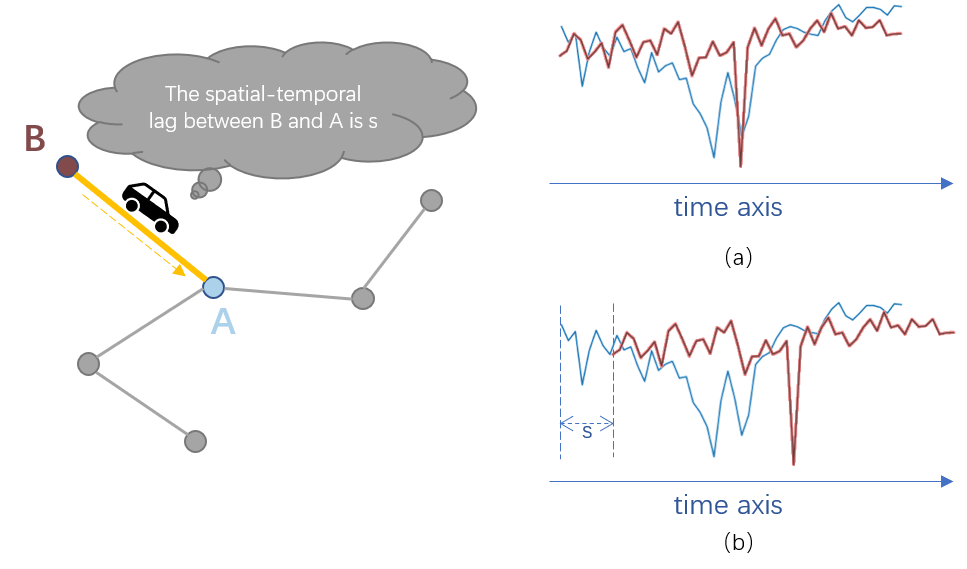}
\caption{Schematic of spatial-temporal alignment. Road network is on left, B is source-node (cause-node), A is target-node (effect-node), and spatial-temporal lag between them is s. (a) Observed time series of A and B. (b) Time series of A and B after spatial-temporal alignment. Blue curve is A’s time series, and dark red is B’s.}
\label{fig8}
\end{figure}

According to the definitions in Section \ref{section3}, for TCR, the source-node is the cause-node, and the target-node is the effect-node. The spatial lag between nodes is the spatial distance between them, and the spatial-temporal lag is the time it takes for GDTi to travel from the cause-node to the effect-node, i.e., the time required for TCR transmission. Since the spatial-temporal lag between nodes is related not only to the spatial distance between nodes but also to the flow velocity of traffic information, which cannot be measured precisely, we use the average velocity of the cause-node as an approximation of the traffic information flow velocity. The algorithm is shown schematically in Figure \ref{fig8}.

\textbf{Algorithm 1 Spatial-temporal alignment. }\label{algorithm1}\emph{Spatial-temporal alignment is the operation that slides the time series of the source-node along the time axis, and the sliding stride is the spatial-temporal lag $s$ between the source-node and target-node. The $s$ is the ratio of the spatial lag between the source-node and target-node to the average traffic velocity of the source-node.}

As illustrated in Figure \ref{fig8}, B is the source-node, and A is the target-node.

\emph{(2)Granger causality test}

After the spatial-temporal alignment of cause-node and effect-node, we can assume that cause-node and effect-node contain the same part of GDTi at the same moment, and we can infer whether there is a transmission of GDTi between them directly from the observed time series. Therefore, we need to answer another important question: how can TCR be detected from long-term observed and dynamically changing time series?

Based on the observation of real-world systems, with the traveling of traffic flow, the traffic information of downstream nodes will contain the traffic information of their upstream nodes, and thus the traffic information of upstream nodes can significantly improve the prediction of downstream nodes. The Granger causality test considers the time-series variables that can significantly improve the prediction after joining as cause variables, while the improved one is the effect variable, which is consistent with our observation. Therefore, we believe that the TCR in the traffic system may be manifested as Granger causality, and using the Granger causality test algorithm, we are able to detect the transmission of traffic information between two nodes, which can express a global and dynamic spatial dependence.

Granger causality analysis determines whether there is a causal relationship between different time-series variables. The basic idea is that if the prediction result using the historical information of both X and Y is better than that of only using the historical information of Y, that is, X helps to explain the future trend of Y, then X is the Granger cause of Y. There are two important hypotheses behind this:

\begin{itemize}
\item Adding information about the cause variable helps recover the information of the effect variable.
\item The time lags need to be modeled when using past data to predict future data.
\end{itemize}

The Granger causality test method established vector autoregressive models for each of the above two predictions as follows.

\begin{equation}
Y_{t+1}=\sum_{j=0}^{m-1} \alpha_j Y_{t-j}+\varepsilon_{Y, t+1} \\
\end{equation}

\begin{equation}
Y_{t+1}=\sum_{j=0}^{m-1} a_j X_{t-j}+\sum_{j=0}^{m-1} b_j Y_{t-j}+\varepsilon_{Y \mid X, t+1}
\end{equation}

where $X_{t-j}$ and $Y_{t-j}$ represent the traffic values of the time series X and Y at time $t-j$  respectively,$\alpha_j$, $a_j$ and $b_j$ represent the regression parameters of the autoregressive model, $\varepsilon_Y$ and $\varepsilon_{Y \mid X}$ represent the residuals of the two autoregressive models respectively, $m$ is the time lag of autoregressive models. If 
$\operatorname{var}\left(\varepsilon_{Y \mid X}\right)< \operatorname{var}\left(\varepsilon_Y\right)$, then it can be determined that the X has a statistically significant Granger causality on Y, expressed as “$X$ Granger-causes $Y$”.
 
\begin{figure}[!t]
\centering
\includegraphics[width=2.5in]{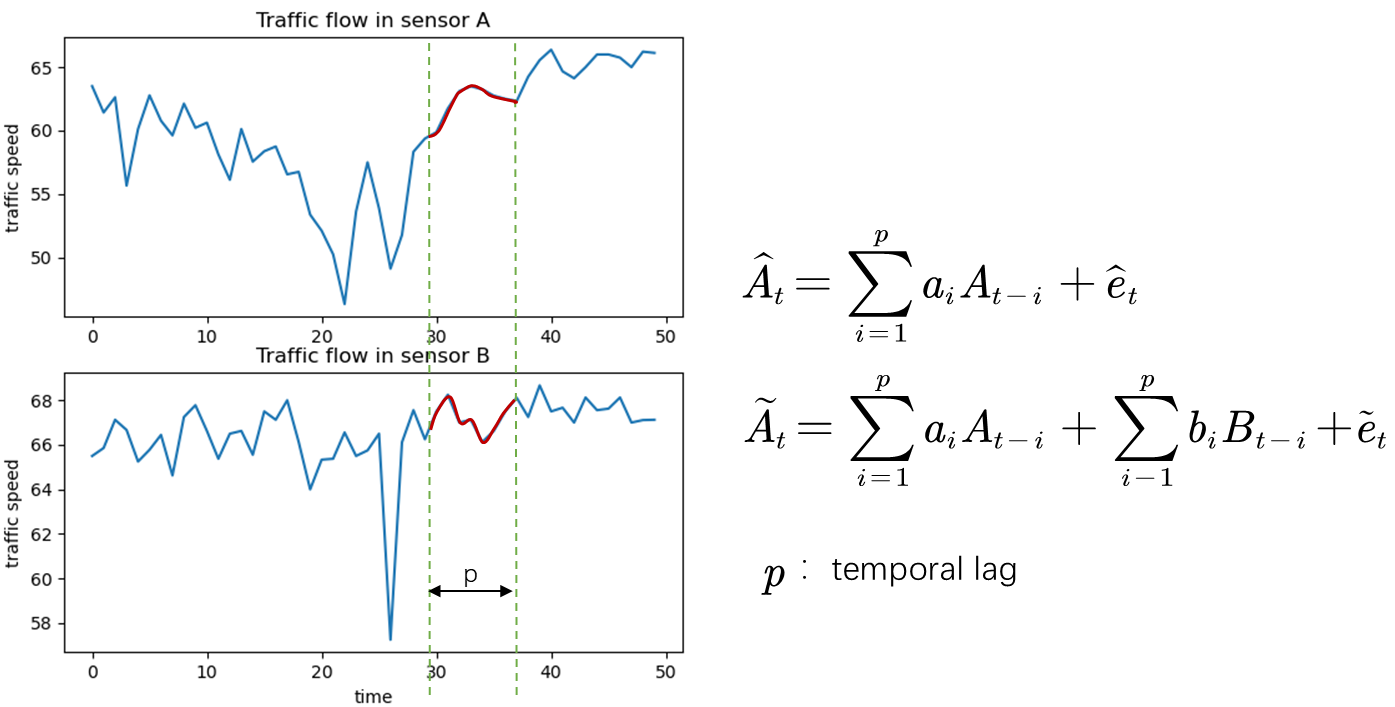}
\caption{Classical Granger causality test. Red curve represents time series used for regression.}
\label{fig9}
\end{figure}

As shown in Figure \ref{fig9}, to perform a Granger causality test on the time series of node A and B, we need to input both time series $X_A=\left\{x_{A_1}, x_{A_2}, \ldots, x_{A_T}\right\}$ and $X_B=\left\{x_{B_1}, x_{B_2}, \ldots, x_{B_T}\right\}$ to be tested. If the returned p-value is below the significance level, the null hypothesis, “$X_B$ does not Granger-cause $X_A$”, is rejected, i.e., the conclusion obtained is that the node B is a Granger cause of the node A. 

\emph{(3)STGC graph}

\begin{figure}[!t]
\centering
\includegraphics[width=2.5in]{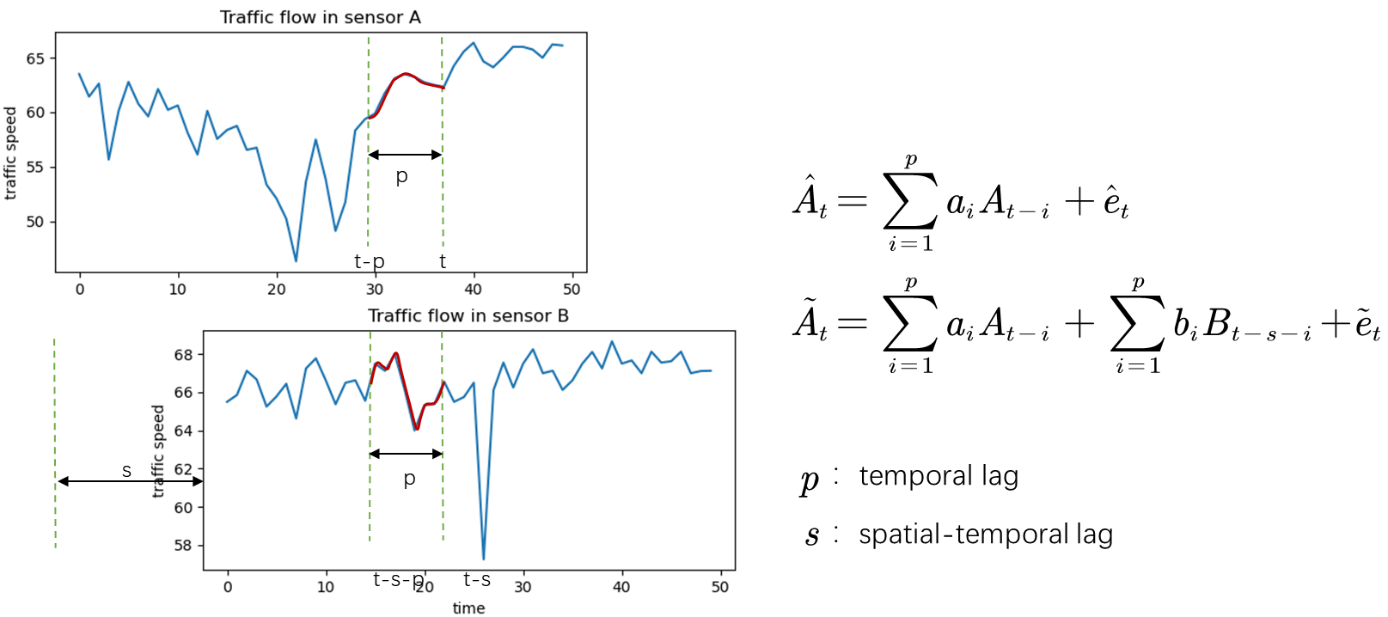}
\caption{Spatial-temporal Granger causality test. Time series were spatiotemporally aligned before being input into Granger causality test module.}
\label{fig10}
\end{figure}

We input the aligned time series of the two nodes into the Granger causality test module, i.e., we use the spatial-temporal Granger causality test method to test the TCR between the two nodes, as shown in Figure \ref{fig10}.

The spatial-temporal Granger causality test is able to detect a Granger causality relationship between two traffic nodes, which can imply global and dynamic spatial dependence. To model this spatial dependence into a graph to input to the GNN for prediction, the following settings are made based on our observations:

\begin{itemize}
    \item \textbf{The reservation of long-range spatial dependence. }To prevent filtering out the long-range spatial dependence, instead of obtaining the input graph by a distance filter or topological adjacency, we model the distance between nodes as a spatial-temporal lag and use the spatial-temporal Granger causality test to detect the spatial dependence under global transmission. It is important to note that we will not use the “cost” offered in the dataset (which implies the spatial distance between different nodes) directly; we calculate the shortest path in the road network between two nodes and obtain the minimum “cost” instead. The reason behind this operation is to reserve as much potential spatial dependence as possible to be detected because the raw dataset does not provide all “cost” between any two nodes, even nodes that are very close. This is also the way to obtain the spatial-temporal lag for long-range spatial dependence because the raw dataset does not provide the “cost” for two nodes that are far away from with a high probability.
    
    \item \textbf{The direction of cause-and-effect.}Spatial-temporal Granger causality, as a manifestation of spatial dependency between nodes, describes a causal relationship underlying traffic information transmission, which can be understood as the tracking of traffic information in the real world. Among them, cause-node corresponds to the source-node of traffic flow, while effect-node corresponds to the target-node of traffic flow. therefore, unlike the spatial graph where the dependence is mostly undirected, the relationships of nodes in the STGC graph are directed, and the direction is from the cause-node to the effect-node. this is not only consistent with the real-world situation that traffic information always flows from source-node to target-node but also consistent with the message passing mechanism of GNN, that is, the message of cause-node can pass to effect-node and update effect-node thus helping effect-node to make better prediction.
\end{itemize}

\subsection{GNN-based traffic prediction}

Eventually, we input the STGC graph into the GNN-based traffic prediction model, which can capture both the temporal dependence and spatial dependence of the traffic road network. The GNN-based traffic prediction model generally combines two modules, one for modeling temporal dependence such as RNN and its variants, and the other for modeling spatial dependence using GNN. Among them, the GNN structure can propagate and aggregate the traffic information between nodes that are spatially dependent, which is a good way to depict the transmission of GDTi. We can evaluate the ability to predict traffic directly by the output result of the GNN-based traffic prediction model.

\section{Experiments and Results}
\label{section5}

\subsection{Dataset}
We conducted experiments on the real-world dataset METR\_LA, which is a Los Angeles highway dataset in the United States and one of the benchmark datasets for traffic prediction. For this dataset, the spatial graph is modeled in the form of a spatial weighting graph, which is later referred to as the spatial distance graph (SD graph), i.e., the spatial distance between nodes is converted into weights using Gaussian filtering.

\emph{(1)Selection of dataset}

Based on \hyperlink{obsevation1}{Observation 1}, we compared the prediction performance of the STGCN on two U.S. highway datasets using self-connected graph (identity matrix as adjacency matrix) and SD graph (spatial weighting matrix as adjacency matrix) as input, the former representing the case of prediction using only the temporal dependence and the latter representing the case of introducing spatial dependence, as shown in Table~\ref{tab1}.

\begin{table}
\begin{center}
\caption{Prediction performance of STGCN on METR\_LA and PEMS\_BAY.}
\label{tab1}
\begin{tabular}{cccc}
\hline
\multirow{2}{*}{Dataset} & \multirow{2}{*}{Horizon} & \multicolumn{2}{c}{MAE/MAPE/RMSE} \\
                   &                    & Spatial weighting matrix         & Identity matrix        \\
\hline
\multirow{3}{*}{METR\_LA} & 15 min  &3.07/7.16\%/6.50 & 3/3.13/7.64\%/6.69\\
                   &30min   &3.55/8.79\%/7.95	&4.12/9.86\%/8.34\\
                   &60min   &4.62/11.37\%/10.05	&4.70/11.41\%/10.31   \\
\hline
\multirow{3}{*}{PEMS\_BAY} & 15 min &1.14/2.30\%/2.28	&1.15/2.36\%/2.32 \\
                   &30min   & 1.41/3.03\%/3.06	&1.41/3.00\%/3.07 \\
                   &60min   & 1.81/4.04\%/3.99	&1.79/4.05\%/4.01 \\
\hline
\end{tabular}
\end{center}
\end{table}

It can be found that on the PEMS\_BAY dataset, the prediction performance of the GNN-based model using a self-connected graph as input is very close to that using the SD graph, while the difference between them on METR\_LA is very significant. Additionally, the overall prediction error value of PEMS\_BAY is smaller. This indicates that the METR\_LA dataset has higher complexity and prediction difficulty and can more significantly compare the effectiveness of different input graphs, so this dataset is chosen for experiments and analysis in this work.

\emph{(2)Basic information of dataset}

\begin{figure}[!t]
\centering
\includegraphics[width=2.5in]{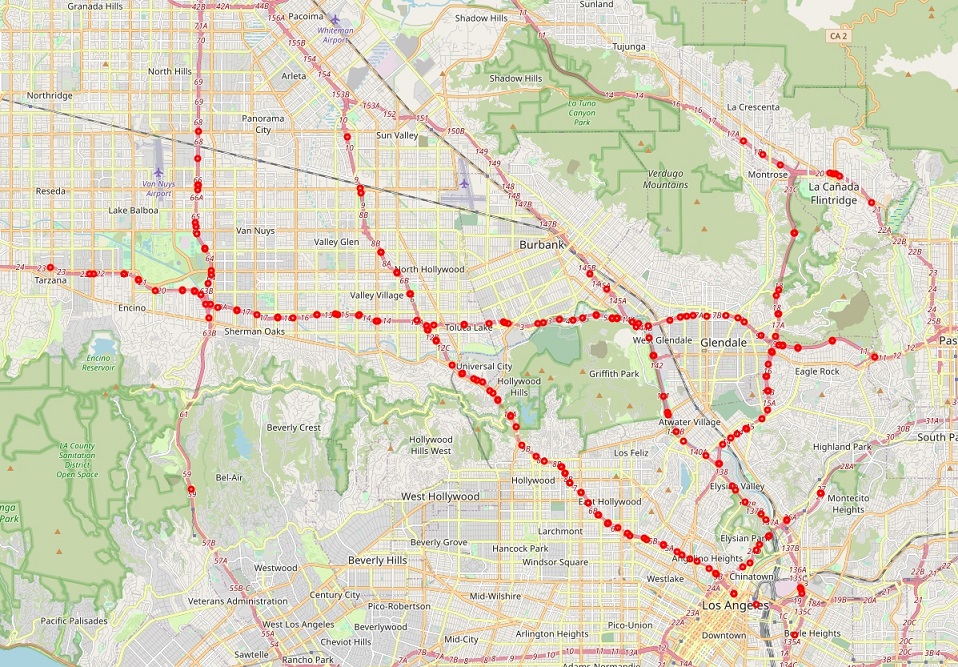}
\caption{Spatial distribution of traffic road network nodes in METR\_LA datasets.}
\label{fig11}
\end{figure}

The basic information of the METR\_LA dataset is listed in Table~\ref{tab2}.The study area and distribution of traffic sensors in the METR\_LA dataset are shown in Figure \ref{fig11}.

\begin{table*}
\begin{center}
\caption{Basic information of METR\_LA dataset.}
\label{tab2}
\begin{tabular}{ccccc}
\hline
Dataset	&Number of sensors	&Duration	&Sampling interval	&Sampling time steps \\
\hline
METR\_LA	&207	&2012/3/1 00:00:00——2012/6/27 23:55:00	&5 min	&34272 \\
\hline
\end{tabular}
\end{center}
\end{table*}

\subsection{Experiments Setting}
To more directly compare the effectiveness of the input graph itself for capturing spatial dependence, we selected three GNN-based models that use the input graph directly instead of performing transformation operations such as the attention mechanism. These models can use the GNN mechanism to aggregate and update messages based on the dependence expressed in the input graph and can directly reflect the prediction improved by neighboring nodes.

\begin{enumerate}
    \item \textbf{T-GCN\cite{19}:} A temporal graph convolution for traffic prediction that combines a graph convolutional network (GCN) and gated recurrent unit (GRU) to capture spatial dependence and temporal dependence.
    \item \textbf{STGCN\cite{21}:}Spatiotemporal graph convolutional networks for traffic forecasting, which builds a model with complete convolutional structures for temporal and spatial prediction.
    \item \textbf{Graph Wavenet\cite{20}:}A deep spatial-temporal graph model for traffic prediction by developing a novel adaptive dependency matrix to work with a spatial distance matrix. To compare our Granger spatial-temporal causal graph with the spatial distance graph, we remove the adaptive dependency matrix as the backbone.
\end{enumerate}

We use these three models as backbone models, i.e., GNN-based traffic prediction models in Figure \ref{fig7}. Apart from this operation, the partitioning of the dataset, the structure of the model and the training settings are kept the same as those of the original model to make a fair and effective evaluation of the effects produced by the method proposed in this paper.

\subsection{Metrics}
To evaluate the effectiveness of this method for traffic prediction tasks, the following metrics are used as a measure of the difference between real traffic data $Y_t$ and predicted data $\hat{Y}_t$.

\begin{itemize}
    \item Mean Absolute Error (MAE)
    \begin{equation}
MAE=\frac{1}{n} \sum_{i=1}^n\left|Y_t-\hat{Y}_t\right|
    \end{equation}
    \item Mean Absolute Percentage Error (MAPE)
    \begin{equation}
MAPE=\frac{1}{n} \sum_{i=1}^n\left|\frac{Y_t-\hat{Y}_t}{Y_t}\right|
    \end{equation}
    \item Root Mean Squared Error (RMSE)
    \begin{equation}
RMSE=\sqrt{\frac{1}{n} \sum_{i=1}^n\left(Y_t-\widehat{Y}_t\right)^2}
    \end{equation}
\end{itemize}

\subsection{Performance Analysis}
We evaluated the performance on prediction horizons of 15 min, 30 min, 45 min, and 60 min. The results are listed in Table~\ref{tab3}, where the backbone using SD graph as input graph uses the original name, while “STGC-” indicates a model using STGC graph as input.

It can be seen that for the same backbone model, the SD graph has better prediction results in all metrics for short-term prediction of 15 min and 30 min, while for the long-term prediction of 45 min and 60 min, the STGC graph has better results. This suggests that spatial dependence constructed from local and static connectivity can capture most of the nodes that interact with the target-node within a short time but seems to fail in long-term prediction. Long-range spatial dependence is needed more for long-term prediction, which is supposed to be detected in the global transmission of traffic information.

The comparison experiments between the STGC graph and the SD graph under the control variable settings have been able to show that the STGC graph is more effective in long-term prediction than the SD graph. To further illustrate that this effectiveness of the STGC graph is self-induced and not just due to the invalidation of the SD graph, we add the results where the input graph is a self-connected graph only (the adjacency matrix is the identity matrix) in comparison. The results are shown in Table~\ref{tab4}, where (-) represents that the input graph is a self-connected graph.

\begin{figure}[!t]
\centering  
\subfigure[]{
\label{fig12a}
\includegraphics[width=2.5in]{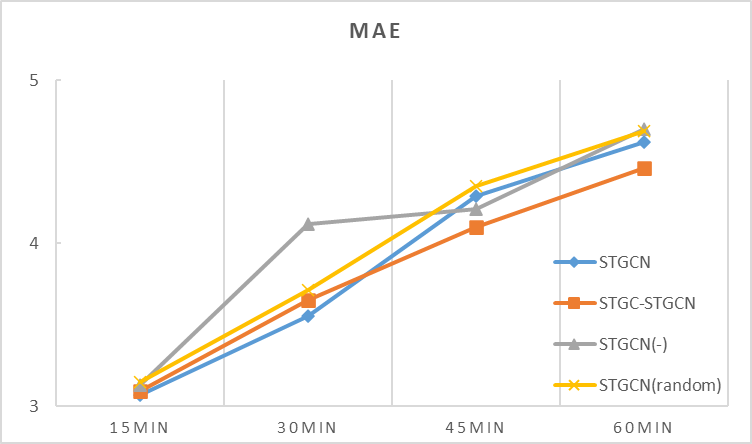}}
\subfigure[]{
\label{fig12b}
\includegraphics[width=2.5in]{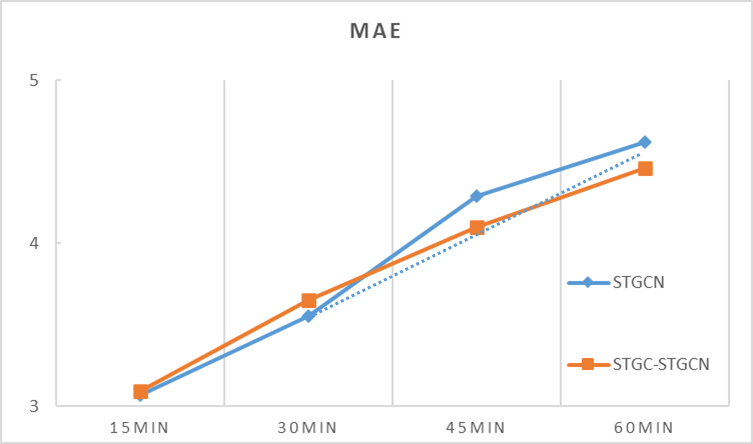}}
\caption{MAEs of different prediction horizons on STGCN model with different input graphs. Lower MAE means more effective input graph. (a) Comparison of performance on STGCN with four types of input graphs. (b) Comparison of performance on STGCN with SD and STGC graphs.}
\label{fig12}
\end{figure}

\begin{table}
\begin{center}
\caption{Prediction results of three backbones on METR\_LA dataset.}
\label{tab3}
\begin{tabular}{ccccc}
\hline
\multirow{2}{*}{Horizon} & \multirow{2}{*}{Model} & \multicolumn{3}{c}{Metrics} \\
                   &                    & MAE     & MAPE(\%)     & RMSE     \\
\hline
\multirow{6}{*}{15 min}  &STGCN             &\textbf{3.07}    &\textbf{7.16} &\textbf{6.50}       \\
                   &STGC-STGCN               &3.09	&7.54	&6.58      \\
                   &T-GCN               &\textbf{3.59}	&9.65	&6.54      \\
                   &STGC-T-GCN              &3.62	&\textbf{9.50}	&\textbf{6.52}     \\
                   &GWNET                    &\textbf{2.53}	&\textbf{6.16}	&\textbf{4.69}     \\
                   &STGC-GWNET   &2.56	&6.39	&4.82  \\
\hline       
\multirow{6}{*}{30 min}  &STGCN             &\textbf{3.55}	&\textbf{8.79}	&\textbf{7.95}     \\
                   &STGC-STGCN               &3.65	&9.02	&8.13      \\
                   &T-GCN               &\textbf{3.86}	&\textbf{10.49}	&7.15      \\
                   &STGC-T-GCN              &3.91	&\textbf{10.49}	&\textbf{7.07}      \\
                   &GWNET                    &\textbf{2.83}	&\textbf{7.29}	&\textbf{5.50}     \\
                   &STGC-GWNET   &\textbf{2.83}	&7.46	&5.53  \\
\hline 
\multirow{6}{*}{45 min}  &STGCN             &4.69	&10.99	&9.41 \\
                   &STGC-STGCN               &\textbf{4.10}	&\textbf{9.99}	&\textbf{9.13}    \\
                   &T-GCN               &4.10	&11.36	&7.62    \\
                   &STGC-T-GCN              &\textbf{4.02}	&\textbf{11.21}	&\textbf{7.45}    \\
                   &GWNET                    &3.06	&8.20	&6.08   \\
                   &STGC-GWNET   &\textbf{3.03}	&\textbf{8.19}	&\textbf{6.01} \\
\hline 
\multirow{6}{*}{60 min}  &STGCN             &4.62	&11.37	&10.05    \\
                   &STGC-STGCN               &\textbf{4.46}	&\textbf{11.24}	&\textbf{10.00}    \\
                   &T-GCN               &4.28	&12.26	&8.08   \\
                   &STGC-T-GCN              &\textbf{4.17}	&\textbf{11.85}	&\textbf{7.82}    \\
                   &GWNET                    & 3.25	&8.78	&6.49   \\
                   &STGC-GWNET   &\textbf{3.17}	&\textbf{8.77}	&\textbf{6.40} \\
\hline 
\end{tabular}
\end{center}
\end{table}

\begin{table}
\begin{center}
\caption{Comparison results between backbones with STGC graph input and self-connected graph input.}
\label{tab4}
\begin{tabular}{ccccc}
\hline
\multirow{2}{*}{Horizon} & \multirow{2}{*}{Model} & \multicolumn{3}{c}{Metrics} \\
                   &                    & MAE     & MAPE(\%)     & RMSE     \\
\hline
\multirow{6}{*}{45 min}  &STGCN(-)             &4.21	&10.41	&9.27 \\
                   &STGC-STGCN               &\textbf{4.10}	&\textbf{9.99}	&\textbf{9.13}\\
                   &T-GCN(-)             &8.55	&30.89	&14.86\\
                   &STGC-T-GCN              &\textbf{4.02}	&\textbf{11.21}	&\textbf{7.45} \\
                   &GWNET(-)                    &3.47	&9.56	&7.07\\
                   &STGC-GWNET   &\textbf{3.03}	&\textbf{8.19}	&\textbf{6.01}\\
\hline 
\multirow{6}{*}{60 min}  &STGCN(-)             &4.70	&11.41	&10.31 \\
                   &STGC-STGCN               &\textbf{4.46}	&\textbf{11.24}	&\textbf{10.00}\\
                   &T-GCN(-)             &8.55	&30.91	&14.87\\
                   &STGC-T-GCN              &\textbf{4.17}	&\textbf{11.85}	&\textbf{7.82} \\
                   &GWNET(-)                    &3.77	&10.67	&7.70\\
                   &STGC-GWNET   &\textbf{3.17}	&\textbf{8.77}	&\textbf{6.40}\\
\hline 
\end{tabular}
\end{center}
\end{table}

In addition, to argue that the validity of such long-term predictions is not due to randomness, we add random groups for comparison. For each prediction horizon, we generated 10 random graphs with the following settings: 1) the sparsity of the random graph is consistent with the STGC graph; 2) for each node, the number of neighboring nodes is consistent with that in the STGC graph; and 3) the IDs of neighboring nodes are randomly generated. The final accuracy is averaged by the evaluation results of 10 random graphs, labeled as STGCN (random) in Figure \ref{fig12}.

Taking the MAE metric as an example, it shows that the predictive power of both STGC-STGCN and STGCN (random) remains relatively steady as the prediction horizon increases, where STGC-STGCN achieves the best results in long-term prediction and STGCN (random) keeps performing poorly, which is not surprising because it randomly introduces information from other nodes. The MAE metric also indicates that the STGC graph’s capability of long-term prediction is not due to random factors such as sparsity but due to the correct capture of spatial dependence. It is noteworthy that the random matrix achieves better results than the identity matrix at a prediction horizon of 30 min and is comparable to the identity matrix at a prediction horizon of 60 min. This illustrates, on the one hand, the limitations of ignoring spatial dependence and, on the other hand, the possibility of achieving better results by chance than by information from itself, even if such luck is not always reliable.

As illustrated in Figure \ref{fig12b}, the original model (STGCN) using the SD graph showed a significant decrease in predictive capability between 30 min and 45 min, while the predictive capability using the STGC graph was much more stable, which also supports our \hyperlink{observation2}{Observation 2}.

\subsection{Visualization}
Although our method outperforms the original model only on the 45-min and 60-min long-term predictions, for the 207 nodes in the METR\_LA dataset, there are nodes that are densely distributed with simple upstream and downstream and thus low prediction difficulty, while there are also nodes located at intersections with higher prediction difficulty. Therefore, the prediction accuracy of individual nodes was calculated and visualized on OpenStreetMap separately. We take the STGCN as an example for analysis in this section.

\begin{figure}[!t]
\centering  
\subfigure[]{
\label{fig13a}
\includegraphics[width=2.5in]{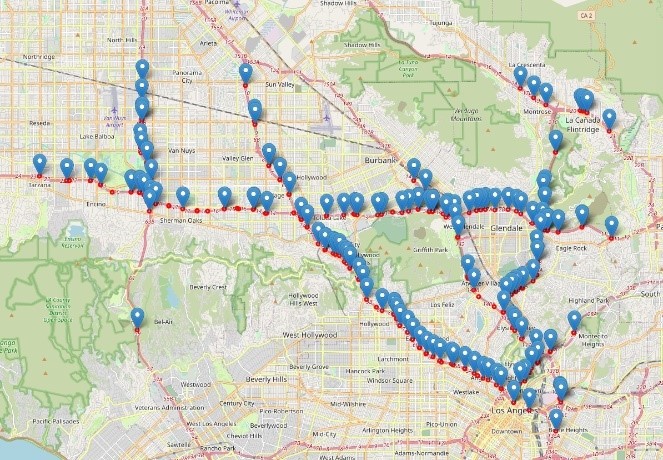}}
\subfigure[]{
\label{fig13b}
\includegraphics[width=2.5in]{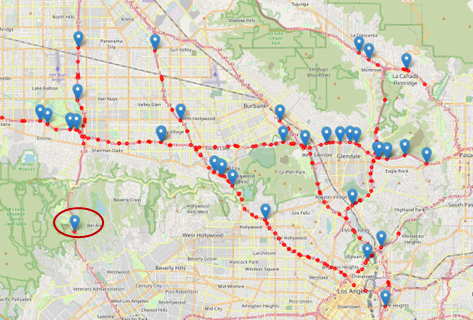}}
\caption{Visualization of nodes that achieve better prediction performance with STGC graph input. (a) Nodes predicted better by STGC graph on horizons of both 45 min and 60 min. (b) Nodes predicted better by STGC graph on all horizons. Circled node is the one that is distant from other nodes.}
\label{fig13}
\end{figure}

\emph{(1) Visualization of the prediction results for the STGC-STGCN}

For the long-term predictions of 45 min and 60 min, our method can achieve better results than the SD graph for most of the nodes (e.g., Figure 13a). For some nodes, our method can achieve better results than the SD graph on all prediction horizons (see Figure 13b), which are mostly located at the boundary and intersection of the road network. We also have better prediction for remote nodes (circle in Figure 13b), which indicates that our method has better results for nodes that are more difficult to predict on all horizons.

\emph{(2) Visualization of STGC-STGCN interpretability}

To interpret the spatial-temporal Granger causality, we detect and compare it with the spatial connectivity inscribed by the SD graph. We visualize the road network and its spatial dependence: the nodes to be predicted are marked with a gray marker, and other nodes are marked with a light gray marker. To visualize the spatial dependence, we used makers with car icons. The cause-node with red marker (with car icon), the effect-node with blue marker (with car icon), and the neighboring nodes in the SD graph with green marker (with car icon).

\begin{figure}[!t]
\centering
\includegraphics[width=2.5in]{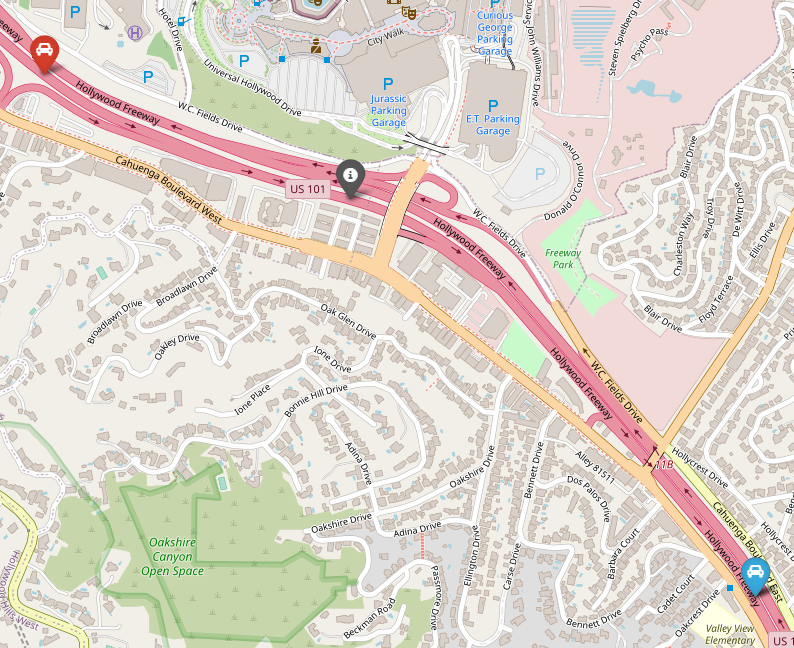}
\caption{Direction of STGC. Red marker is cause-node, and blue marker is effect-node we detected for node to predict (light gray marker). White dotted line is trajectory we inferred from cause-and-effect.}
\label{fig14}
\end{figure}

\textbf{Direction of STGC. }The STGC graph describes a kind of spatial dependence caused by global traffic information transmission, which corresponds to the upstream and downstream relationship of traffic flow in the real world. The STGC graph describes a kind of spatial dependence caused by global traffic information transmission, which corresponds to the upstream and downstream of traffic flow in the real world. By examining the detected causal relationship with the directions marked on the highway by OpenStreetMap, the effect-node should be in the downstream direction of the cause-node. As shown in Figure \ref{fig14}, it can be found that the cause (upstream) and effect (downstream) of our detection are consistent with the direction of the highway marked on the map.

The local connectivity constructed in the SD graph is a static relationship generated from the invariant spatial structure of the road network (node location distribution, road network topology, etc.), which mostly has no direction and is unable to portray the interaction caused by dynamic traffic flow. However, driving on a highway is constrained by traffic rules, and it is difficult to have direct interactions between nodes on roads in different driving directions even if they are close to each other. Spatial dependence is likely to be wrong when determined only by spatial distance but without considering the possibility of actual interaction.

\begin{figure}[!t]
\centering
\includegraphics[width=2.5in]{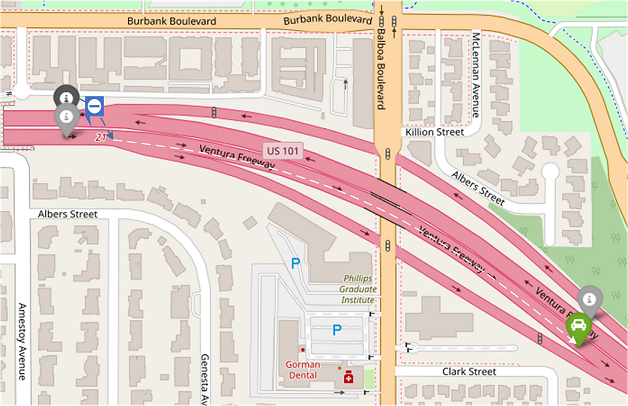}
\caption{Misjudgment of SD graph. Green marker (with car icon) is neighbor of gray marker in SD graph, but they are not directly reachable for each other.}
\label{fig15}
\end{figure}

As shown in Figure \ref{fig15}, two nodes that are spatially dependent in the SD graph do not actually reach each other, and traffic flow interactions cannot be generated between them. Considering the possibility of actual interactions, it is likely to introduce incorrect spatial dependence.

\begin{figure}[!t]
\centering
\includegraphics[width=2.5in]{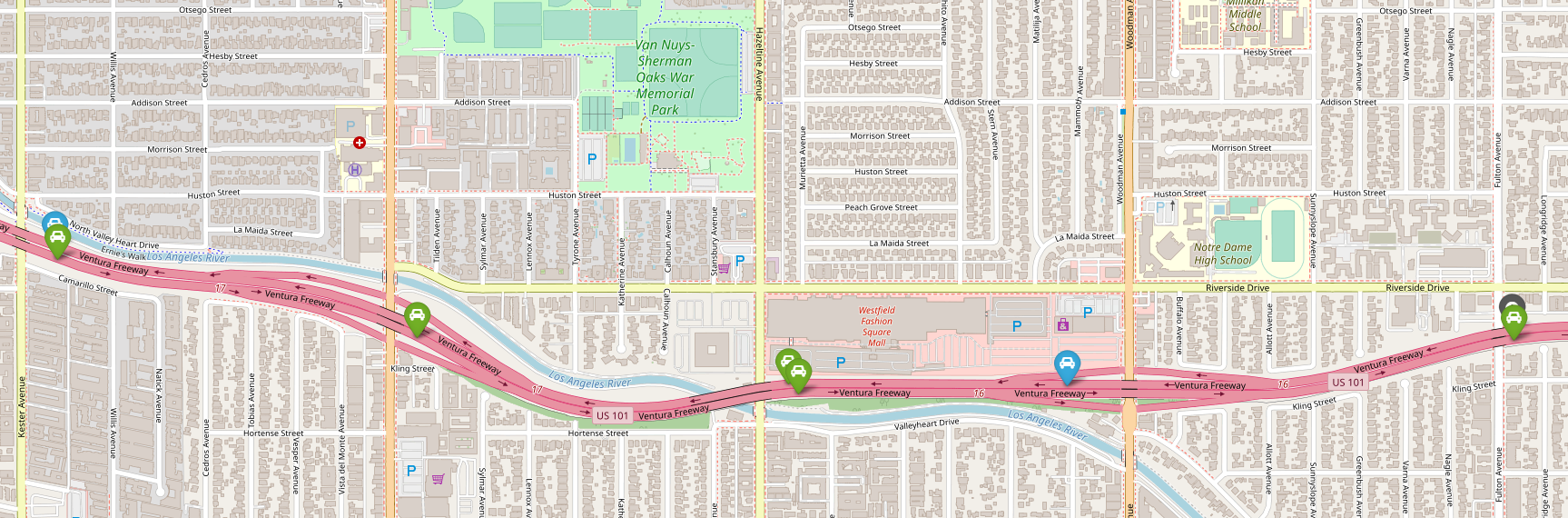}
\caption{Judgment of direction made by STGC graph. Green markers (with car icon) are neighbors of gray marker in SD graph. Some may appear on road in opposite direction and are not actually reachable.}
\label{fig16}
\end{figure}

Spatial-temporal Granger causality in a sense can avoid the introduction of incorrect information by detecting the global transmission of traffic information between two nodes. As shown in Figure \ref{fig16}, the effect-node of the node to predict always appears downstream, while the spatially connected node may appear on the road in the opposite direction, which cannot be either upstream or downstream.

\begin{figure}[!t]
\centering
\includegraphics[width=2.5in]{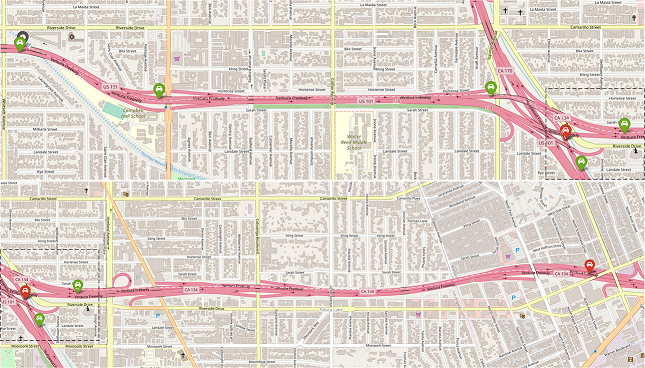}
\caption{Long-range dependence captured by STGC graph. Upper and lower pictures can be combined according to dashed box. Node to be predicted is at left of upper picture, and long-range spatial dependence that can be captured by spatial-temporal Granger causality can extend to right of lower picture.}
\label{fig17}
\end{figure}

\textbf{STGC captures long-range spatial dependence. }The STGC graph can model long-range spatial dependence, while the SD graph can only model local connected nodes, which will not spend a longer time interacting with the node to be predicted. As shown in Figure \ref{fig17}, the spatial dependence between the gray marker (with an icon of a car) in the upper picture and the red marker (with an icon of a car) in the lower picture can be detected by the STGC graph.

\section{Discussion}
\label{section6}

In this section, we analyze the possible reasons why spatial-temporal Granger causality can outperform the SD graph at 45 min and longer horizons. We also summarize the shortcomings of our method and suggest possible future solutions.

\textbf{Spatial distance does not mean everything. }In many traffic prediction benchmark datasets, such as METR\_LA used in this paper, the spatial distance between nodes is not the actual geographical distance between them but the distance traveling from one node to another along the road network, and this distance needs to be actually measured. Therefore, in the raw dataset, there may be missing data, i.e., the distance values of two neighboring nodes are missing. However, the SD graph is constructed considering only the distances recorded in the dataset, so the situation in Figure 18 may occur, where the closer nodes are not connected in the SD graph, but instead the distant nodes are connected even if the closer nodes are reachable to each other.

In addition, in a traffic system with obvious physical constraints between roads with different directions such as a highway, two nodes may take a high cost to connect even though they are very close in space, and if these two nodes are determined to be spatially dependent, then they may bring poor results for prediction. Similar to the scenario in Figure 19, in the SD graph generated by spatial distance filtering, two nodes on roads with opposite directions are judged as neighbors. Our method further determines this spatial dependence by detecting the transmission of traffic information, which can avoid this negative influence to a certain extent.

\textbf{Time to travel across the road network. }We obtained statistics on the spatial-temporal lag of each pair of nodes in the road network and found an interesting phenomenon. In the ideal case (i.e., the interaction between nodes is planned by the shortest path and kept at the average flow rate of the source-node for smooth transmission), most nodes can complete the interaction within 30 min (spatial-temporal lag <= 6), and the nodes farthest apart can complete the interaction within 45 min (spatial-temporal lag<=9), which is exactly the capability boundary of the SD graph. That is, when the prediction horizon > 45 min, a large amount of traffic flow may have already left the system, and the new flow entering the system is unknown; therefore, the overall flow in the system is not guaranteed to be constant. In this case, the uncertainty of traffic information transmission further becomes higher, and the spatial dependence to be captured by the prediction is still long-range dependence, but it can no longer be inferred from the spatial location alone, and a stable association must be mined from the observed data. Thus, we hypothesize that local and static connectivity is insufficient to cope with this uncertainty when the prediction duration exceeds the maximum traveling time in a traffic system. The spatial-temporal Granger causality test method we use essentially mines a stable spatial dependence from time series that are long-term observed and dynamically changing and is able to address the problem of long-term prediction to some extent. How exactly this uncertainty is modeled and how the transmission of traffic flow is tracked are questions that can be considered and solved in the future.

\textbf{Spatial dependence is dynamically changing or invariant? } Traffic flow is actually observed at the macro level generated by a large number of microlevel vehicles traveling in the road network, which have very high uncertainty in their trajectories but exhibit some patterns at the macro level. The spatial pattern is recognized by spatial dependence. The construction of an SD graph essentially establishes spatial dependence from a static road network with a constant spatial structure, and the assumption behind it is that the closer they are, the stronger the spatial dependency, so this spatial dependence is actually static and local. However, as we observe, traffic time series are actually dynamic and complex over time and space and potentially influenced by factors such as the environment and society. Static spatial dependence is no longer sufficient to describe the changing spatial dependence of the road network. We assume that although the spatial dependence of the road network is dynamically changing, there will be an equilibrium, i.e., there is a stable spatial dependence, and the changing spatial dependence always approaches this steady state. Thus, this steady state can explain the spatial dependence of the road network to the greatest extent and help to predict it. Our method attempts to approximate this stable spatial dependence. Therefore, in some time windows or prediction horizons, our prediction result may deviate from the real situation. In the future, we can choose to divide the input into time periods and detect an STGC graph for each time period to realize the dynamic modeling of spatial dependence, which may work better for complex traffic prediction problems.

\textbf{What is the role of temporal graphs? } There is another class of GNN-based traffic prediction models that combine temporal graphs with spatial graphs as input. In this type of model, the temporal graph is often chosen to connect nodes with higher similarity in time series, which can compensate to some extent for the spatial graph that only uses static and local connectivity. However, this type of model eventually ensures the effect in decision-making by operations such as cropping or maxpooling and actually still relies mainly on spatial graphs. In this paper, we do not compare and improve such models because our approach also detects relationships from observed time series, which may create redundancy with the information contained in the temporal graph. We believe that the temporal graph actually models the effect of spatial dependence on the time axis and can help model spatial dependence more accurately, but how to play its role is an open question to be discussed.

\section{Conclusions}
\label{section7}

In this paper, we proposed a spatial-temporal Granger causality to model spatial dependence in the road network. This causality can capture the transmission of GDTi. Unlike the local and static connectivity generated from the spatial distribution of the road network, the spatial-temporal Granger causality we proposed can capture the long-range dependence and approximate the stable causal relationship of dynamic traffic flow transmission from the long-term observed traffic flow and thus performs better in 45-min and 60-min long-term predictions. Specifically, we proposed a TCR hypothesis and a spatial-temporal Granger causality test method to detect causal relationships.

The causal relationship we eventually obtained is directional and can match well with the source-node and target-node of a real-world traffic system. Experiments on the METR\_LA dataset showed that our method can outperform the original model on three GNN-based traffic prediction models at 45 min and 60 min. At the prediction level of individual nodes, our model can improve the prediction at all horizons for intersectional, boundary, and distant nodes. The upstream and downstream traffic information transferred at intersectional nodes are complex, and boundary nodes tend to depend only on upstream or downstream nodes, while remote nodes rely more on the capture of long-range dependence. This also illustrates the effectiveness of our method for GDTi transfer capture compared to a spatial graph.

\ifCLASSOPTIONcaptionsoff
  \newpage
\fi



\bibliographystyle{IEEEtran}
\bibliography{IEEEabrv,reference}
\end{document}